%% file: main.tex
\documentclass[sigconf]{acmart}

\usepackage{booktabs} 
\usepackage[utf8]{inputenc}
\usepackage[T1]{fontenc}
\usepackage[linesnumbered,ruled]{algorithm2e}
\usepackage{color}
\usepackage{amssymb}
\usepackage{pifont}
\usepackage{booktabs}
\usepackage{tabularx} 
\usepackage{algorithmicx,algpseudocode}
\usepackage{ifthen}
\usepackage{amsmath}
\usepackage{array}
\usepackage{tabu}
\usepackage{epsfig}
\usepackage{amssymb}
\usepackage{amsmath}
\usepackage{amsfonts}
\usepackage{bbding}
\usepackage{float}
\usepackage{fancyhdr}
\usepackage{textcomp}
\usepackage{gensymb}
\usepackage{enumitem}

\usepackage{graphicx}
\usepackage{caption}
\usepackage{subcaption}

\newcommand{\sys}{\textsc{DriveContext}}

\newcommand{\thickhline}{%
    \noalign {\ifnum 0=`}\fi \hrule height 1pt
    \futurelet \reserved@a \@xhline
}

\usepackage{silence}

\usepackage{color, colortbl}
\definecolor{Gray}{gray}{0.9}

\usepackage{setspace}

\usepackage[T1]{fontenc}

\copyrightyear{2017} 
\acmYear{2017} 
\setcopyright{rightsretained} 
\acmConference{SIGSPATIAL'17}{November 7--10, 2017}{Los Angeles Area, CA, USA}
\acmDOI{10.1145/3139958.3139992}
\acmISBN{978-1-4503-5490-5/17/11}

\begin{document}
\setstretch{.9}
\setlength{\pdfpagewidth}{8.5in}
\setlength{\pdfpageheight}{11in}

\vspace{-0.2in}
\title{Characterizing Driving Context from Driver Behavior}
\author{Sobhan Moosavi$^{\dag}$, Behrooz Omidvar-Tehrani$^{\dag}$, R. Bruce Craig$^{\S}$, Arnab Nandi$^{\dag}$, Rajiv Ramnath$^{\dag}$}
\affiliation{
  \institution{$^{\dag}$The Ohio State University, $^{\S}$Nationwide Insurance}
  }
\email{{moosavinejaddaryakenari.1,omidvar-tehrani.1,nandi.9,ramnath.6}@osu.edu, craigr2@nationwide.com}

\begin{abstract}
Because of the increasing availability of spatiotemporal data, a variety of data-analytic applications have become possible.
Characterizing {\em driving context}, where context may be thought of as a combination of {\em location} and {\em time}, is a new challenging application. An example of such a characterization is finding the {\em correlation} between driving behavior and traffic conditions.
This contextual information enables analysts to validate observation-based hypotheses about the driving of an individual. 
In this paper, we present \textsc{DriveContext}, a novel framework to find the characteristics of a context, by extracting significant driving patterns (e.g., a slow-down), and then identifying the set of potential causes behind patterns (e.g., traffic congestion).
Our experimental results confirm the feasibility of the framework in identifying meaningful driving patterns, with improvements in comparison with the state-of-the-art. We also demonstrate how the framework derives interesting characteristics for different contexts, through real-world examples.
\end{abstract}

%
%



\maketitle

\input{introduction}
\input{ProblemStatement}
\input{DriveContext}
\input{results}
\input{literature}
\input{conc}

\bibliographystyle{ACM-Reference-Format}
\bibliography{main} 

\end{document}

%% file: introduction.tex
\section{Introduction}
\label{sec:intro}
The amount and availability of spatiotemporal data has drastically increased thanks to the ubiquity of sensors in various applications and high-capacity data centers that can store and serve up this data. 
Transportation data, like New York Taxicab \cite{nycTaxi} data, is an example of such spatiotemporal data.
Given the availability of these large transportation data, various analysis applications have been developed to gain insights from this data. 
{\em Characterizing driving context} is a new application area which we introduce in this paper. A {\em context} can be described as a combination of {\em location} (e.g., Interstate-90) and {\em time} (e.g., weekdays between 3pm to 7pm). A {\em characteristic} for a context can be identified as a correlation between driving behavior and an environmental effect (e.g., traffic congestion).
By having information about different driving contexts, one can validate hypotheses about behavior of an individual within a context, and also provide feedback to drivers in order to help them to improve their skills. The former is related to {\em usage-based insurance} \cite{ubi2017} and the latter to {\em driver coaching} \cite{stanton2007changing}. 

In this paper, we address the problem of {\em Characterizing driving context}. Our characterization is based on the observed behavior of drivers and exploiting complementary sources of spatiotemporal data to analyze this behavior. 
We define ``driver behavior'' in terms of meaningful {\em driving patterns} (e.g., {\em turn}, {\em speed-up}, {\em slow-down}, etc.).
In addition, we try to explore the {\em causes} which underlie a specific pattern within a context by conducting analyses across several spatiotemporal data sources (traffic data, road features, etc.).
The cause behind a transition between patterns, hence introducing a new pattern, can be {\em extrinsic} (e.g., an accident, a traffic signal, a traffic congestion) or {\em intrinsic} (e.g., driver-generated distraction,  or the personality of the driver).

Characterizing driving context is a challenging problem, due to the lack of supervision on drivers and/or environment. Unlike previous studies in fully monitored environment (that use, for examples, cameras placed inside the car) \cite{liu2001modeling,sathyanarayana2008driver}, we focus on a dataset with only externally visible phenomena (e.g., vehicle's speed) with no monitoring on drivers and environment. Given these limitations, the goal is to learn driving context based simply on the behavior of drivers.
In this paper, we introduce \sys, a framework to efficiently characterize driving contexts. This framework consists of two components, {\sc dSegment} and {\sc dDescribe}. The former applies a behavior-based trajectory segmentation approach to find meaningful driving patterns within a trajectory.
The latter reveals the extrinsic causes for each driving pattern. We apply \sys\ on a real-world dataset of car trajectories to derive interesting characteristics for different contexts.

%% file: ProblemStatement.tex
\vspace{-0.12in}
\section{Problem Statement}
\label{sec:problem_statement}
Assume we are given a set ${\mathcal T} = \{ T_1, T_2, \dots, T_n \}$ of trajectories.
For $1 \leq i \leq n$, we define trajectory $T_i = \langle p_{i1}, p_{i2}, \dots, p_{im} \rangle$, where $p_{ij} \in T_i$ is a data point of the form $\{t, lat, lng, s, a, h\}$ which captures a vehicle's status at time $t$ as its latitude and longitude are $\langle lat, lng \rangle$, with speed $s$ (km/h), acceleration $a$ ($m/s^2$), and heading $h$ (degrees).
We study the ``discovery of driving context'' in terms of two sub-problems: {\em Segmentation} and {\em Causality Analysis}. 

A segmentation of a trajectory $T$ into $k$ segments, denoted as $seg^k_T$, is to find a set of cutting indexes $seg^k_T = \langle I_1, I_2 \dots, I_k \rangle$ to mark the end points of the segments. Thus, we can define a set of cutting data points for the segmented trajectory $T$ as $\langle p_{I_1}, p_{I_2} \dots, p_{I_k} \rangle$. Note that $p_{I_k}=T_{|T|}$. 
Each segment represents a {\em driving pattern} and each cutting point $p_{I_i}, I_i \in seg^k_T$, represents a {\em transition between patterns}.

We assume the existence of a segment is potentially relevant to extrinsic or intrinsic causes. In this work, the focus is on extrinsic causes that we refer to as $events$. We keep track of events in an event database ${\mathcal E}$ of the form $e = \langle t,$ $lat, lng,$ $type\rangle$, where each event $e \in {\mathcal E}$ occurs in time $t$, in a geographical area whose center is $\langle lat, lng \rangle$ of type $type$. 
Given a set of cutting points $\langle p_{I_1}, p_{I_2} \dots, p_{I_k} \rangle$ by segmenting trajectory $T$, and the database ${\mathcal E}$ of events, the second sub-problem (i.e., causality analysis) is one of finding if, and to what extent, each cutting point $p_{I_i}$, $1\leq i\leq k$, is related to (or caused by) an event $e \in {\mathcal E}$.

%% file: DriveContext.tex
\section{The \sys\ Framework}
\label{sec:DriveContext}
Figure \ref{fig:dc_framework} depicts the overall process of \sys\, where it consists of two major components, {\sc dSegment} and {\sc dDescribe}.

\vspace{-0.05in}
\subsection{{\textsc dSegment} Component}
\input{dSegment}

\vspace{-0.05in}
\subsection{{\textsc dDescribe} Component}
\input{dDescribe}

%% file: dSegment.tex
\label{sec:dSegment}

\begin{figure}[t]
  \centering
  \includegraphics[scale=0.35]{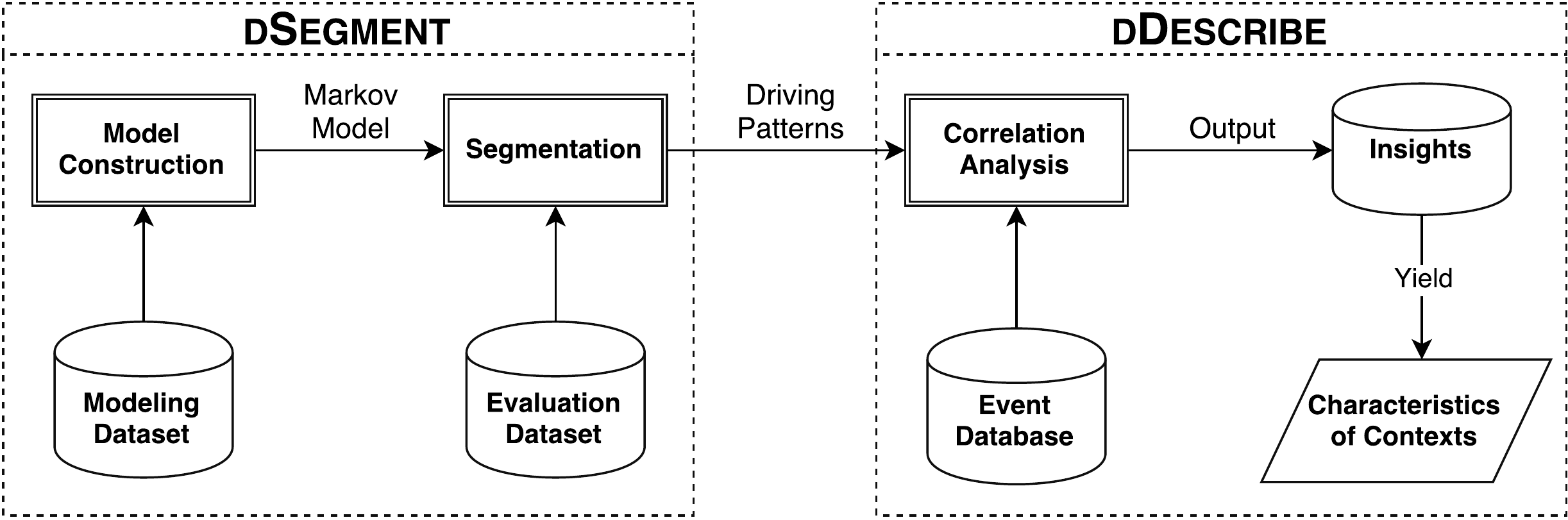}
  \vspace{-0.19in}
  \caption{\small The overall process of \sys\ framework}
  \label{fig:dc_framework}
  \vspace{-0.1in}
\end{figure}

{\sc dSegment} is an adaptation of our previously proposed approach~\cite{moosavi2016discovery} to wisely partition a trajectory based on behavior of driver, such that each resulting segment corresponds to a meaningful driving pattern (e.g., turn, speed-up, etc.). 
A summary of fundamental parts of {\sc dSegment} is provided as follows.

\vspace{2pt}
\noindent {\bf Dataset Preprocessing}: This step includes several data scrubbing tasks such as removing records with noisy GPS data, rounding values of acceleration and heading to simplify the model, and using {\em change of heading} instead of absolute heading to reflect changes clearly.

\vspace{2pt}
\noindent \textbf{Markov Model Creation}: We model behavior of a population of drivers in terms of a Markov Model, a finite state machine whose edges are labelled with the probability of transition from one {\em driving state} to another. In order to deal with sparsity of the model and avoid overfitting, we adapt a regularization technique known as the {\em Wedding Cake} technique \cite{krumm2006predestination}. A portion of the model graph is shown in Figure~\ref{fig:sample_markov}. 

\vspace{2pt}
\noindent \textbf{Trajectory Transformation}: The next step is to transform an input trajectory into a {\em signal} in a new space called Probabilistic Movement Dissimilarity (PMD) space. The generated signal for a trajectory shows how unlikely the behavior of a driver is during different moments of a trip. Algorithm~\ref{algo:transformation} summarizes the transformation process.

\vspace{2pt}
\noindent \textbf{Trajectory Segmentation}: Finally, we apply a dynamic programming approach \cite{han2004optimal} to segment the generated signal.

%% file: dDescribe.tex
\label{sec:dDescribe}
{\sc dDescribe} analyzes the extracted segments (which are essentially driving patterns) to explore the underlying causes behind these patterns. 
This step identifies the characteristics for a given context. 
The existence of a driving pattern is potentially related to extrinsic or intrinsic causes.
Recall that segments of a trajectory $T$ are represented in form of cutting points $\langle p_{I_1}, p_{I_2} \dots, p_{I_k} \rangle$. 
Having a database of events ${\mathcal E}$ and a cutting point $p_{I_i}$, $1\leq i\leq k$, the goal is to find whether $p_{I_i}$ is {\em related} to an event $e \in {\mathcal E}$ or not. 
If we find that $p_{I_i}$ is related to $e$, then the segment which starts at the cutting index $(I_i + 1)$ is potentially caused by event $e$. 
We define the relevancy relationship between a cutting point $p$ and an event $e$ based on the type of the event. In this study, we consider two types of events: {\em physical fact} and {\em temporal-physical event}. We measure the relevancy for each type of event as described below. 

\vspace{2pt}
\noindent{\bf Physical Fact.} An example is the presence of a traffic signal. In such a case, the relevancy can be measured as the distance between the locations of cutting point $p$ and event $e$. We then say $p$ and $e$ are correlated if their locations are within a specified distance threshold.

\vspace{2pt}
\noindent{\bf Temporal-Physical Event.} An example is the existence of a traffic congestion in a specific location during a time interval. In this case, we say  $p$ and $e$ are correlated if the two following conditions are satisfied: $(i)$ the time of the trajectory $T$, where $p \in T$, overlaps with the time interval of the event $e$, and $(ii)$ the distance between locations of $p$ and $e$ are lower than a threshold.

%% file: results.tex
\vspace{-0.1in}
\section{Evaluation}
\label{sec:eval}

We first describe the datasets we used in our study. Then, we evaluate {\sc dSegment} with respect to a ground-truth dataset. Next, we apply {\sc dSegment} on a real-word dataset of car trajectories and conduct causality analysis using {\sc dDescribe}. 

\begin{figure}[t]
  \centering
  \includegraphics[scale=0.23]{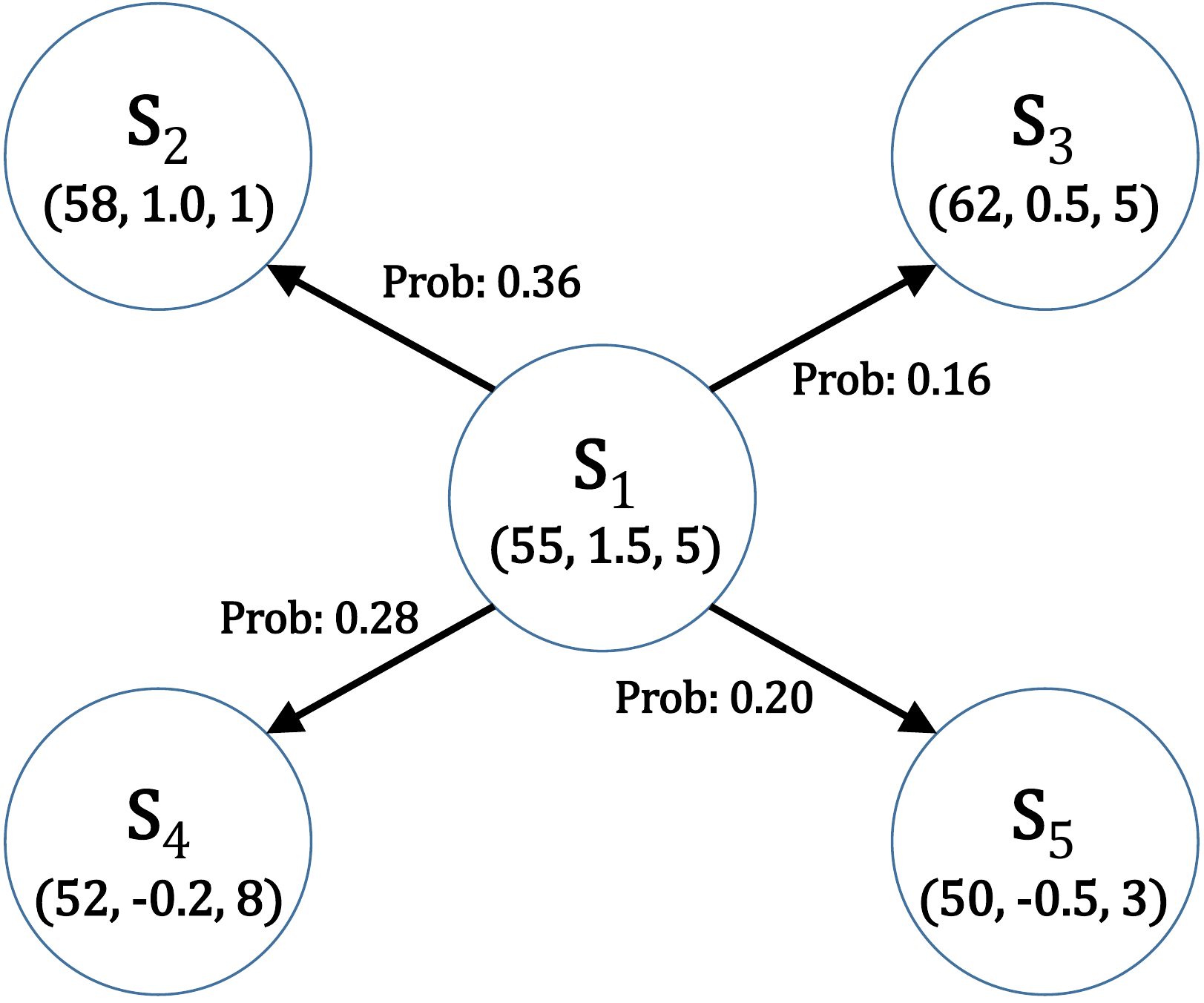}
  \vspace{-0.08in}
  \caption{\small A portion of the Markov Model. Each state is a triple of Speed ($km/h$), Acceleration ($m{^2}/s$), and Change of Heading. Probability of transition is shown on each edge.}
  \label{fig:sample_markov}
\end{figure}

\begin{algorithm}[t]
\footnotesize
\DontPrintSemicolon
\KwIn{Trajectory $T$, Markov Model $M$}
$Signal_T \gets \langle\rangle$ \Comment{\footnotesize This is PMD signal for $T$}\;
\For {$i=1$  \textbf{to}  $|T|-1$}
{
    $\phi \gets$ State in $M$ correspond to $p_i$\;
    $\phi' \gets$ State in $M$ correspond to $p_{i+1}$\;
    $prob_{\phi\rightarrow \phi'}\gets$ Probability of transition from $\phi$ to $\phi'$ in $M$\;
    $R \gets$ Set of transitions from state $\phi$ in $M$\;
    $v = 0$\;
    \For{$r \in R$}  
        {
            $prob_{\phi\rightarrow r} \gets$ Probability of transition from $\phi$ to $r$ in $M$\;
            $v$ += $EuclideanDistance(\phi',r)\times prob_{\phi\rightarrow r}$\;
        }
    $v = \frac{v}{|R|}$\;
    $Signal_T \gets Signal_T \textasciicircum v$ \Comment{\footnotesize Append $Signal_T$ by value $v$}\;
}
\KwOut{$Signal_T$ for trajectory $T$}
\caption{Trajectory Transformation}
\label{algo:transformation}
\end{algorithm}

\vspace{-0.1in}
\subsection{Datasets}
\input{datasets}

\vspace{-0.1in}
\subsection{{\textsc dSegment} Evaluation}
\input{SegmentationResults}

\vspace{-0.1in}
\subsection{{\textsc dDescribe} Evaluation}
\input{CausalityResults}


%% file: datasets.tex
\label{sec:dataset}
We used four different sets of spatiotemporal data sources to build and evaluate components of \sys. 

\vspace{2pt}
\noindent \textbf{Dataset of Annotated Car Trajectories (DACT):} We have constructed a dataset of annotated car trajectories to evaluate {\sc dSegment}~\cite{moosavi2017dact}. 
DACT consists of two sets of annotations for each trajectory, one that assumes ``flexible'' constraints to identify segment borders (Easy-Aggregation), and the other that uses ``strict'' constraints (Strict-Aggregation). DACT includes $50$ trajectories covering about $13.3$ hours of driving data. 

\vspace{2pt}
\noindent \textbf{Nationwide Trajectories:} We use a rich dataset of trajectories provided by an insurance company in the state of Ohio, in the United States. To our knowledge, this dataset, which have named Nationwide Trajectories, is one of the few large scale datasets with driving data for personal vehicles (as opposed to commercial transportation vehicles). The dataset contains 83,406 trajectories and covers about 20,689 hours of driving data. 
We divided the Nationwide Trajectories into two sets, {\em train} and {\em test}, with 81,895 and 1,421 trajectories respectively. The former is used to build the {\sc dSegment} model, and the latter is used to evaluate {\sc dDescribe}. The test set contains sampled data for 5 popular routes in the city of Columbus Ohio (Table \ref{tab:segmentation_result}).  

\vspace{2pt}
\noindent \textbf{Physical Facts:} Physical facts consist of annotations for routes in the test set of Nationwide dataset and are drawn from two different sources of data, i.e., Open Street Map (OSM)\footnote{\scriptsize www.openstreetmap.org} (with annotations like {\em exit}, {\em merge}, and {\em bridge}) and Hand-Curated Annotations (HCA) usig Google Street View to complement the former. The set of physical facts contains 1,825 annotations from OSM and 95 from HCA.

\vspace{2pt}
\noindent \textbf{Temporal-Physical Events:} We employ Bing Traffic API\footnote{\scriptsize https://msdn.microsoft.com/en-us/library/hh441725.aspx} and Map Quest Traffic API\footnote{\scriptsize https://developer.mapquest.com/products/traffic} to collect temporal-physical events such as real-time traffic reports. The dataset contains 1,688 records from Bing and 4495 records from MapQuest for routes in the  test set. 

%% file: SegmentationResults.tex
\label{sec:segmentation_results}

\begin{table}[t]
    \footnotesize
    \caption{\small Summary of evaluation set and segmentation outcome}\vspace{-0.01\textwidth}
    \centering
    \begin{tabular}{| c | c | c | c | c |}
        \hline
        \rowcolor{Gray}
        \textbf{Route} & \textbf{\begin{tabular}{@{}c@{}} Route \\ Length\end{tabular}} &
        \textbf{\begin{tabular}{@{}c@{}} Number of \\ Trajectories\end{tabular}} & \textbf{\begin{tabular}{@{}c@{}} Avg. Trajectory \\ Length (secs) \end{tabular}} & \textbf{\begin{tabular}{@{}c@{}} Avg. Number \\ of Segments \end{tabular}} \\ 
        \hline
        Interstate-70 & 6.7 km & 535 & 333 & 4\\
        \hline
        Interstate-71 & 13.8 km & 438 & 546 & 5\\
        \hline
        Interstate-270 & 12.3 km & 195 & 444 & 3.3\\
        \hline
        Interstate-670 & 7.0 km & 131 & 359 & 4.2\\
        \hline
        315 Freeway & 14.6 km & 120 & 703 & 8.9\\
        \hline
    \end{tabular}
    \label{tab:segmentation_result}
    \vspace{-0.15in}
\end{table}

We use the training set of Nationwide trajectories to create the Markov model. 
The regularized Markov model consists of 47,495 states and 5.8 million transitions between states.
In order to evaluate our segmentation approach, we use the DACT annotation sets \cite{moosavi2017dact}. For comparison purposes, we use following four baselines: $(i)$ {\em Stable Criteria}, where a set of spatiotemporal heuristics (criteria) are used for segmentation \cite{buchin2010algorithmic,alewijnse2014framework}; $(ii)$ {\em Point of Change Detection}, where we employ the change point detection approach in \cite{liu2013change} for comparison; $(iii)$ {\em Equal Length}, where we first assume all trajectories have the same number of segments ($\eta$) and then divide a trajectory to $\eta$ equal size segments; and $(iv)$ {\em Random}, where we find $\eta$ segment borders at random to form $\eta$ segments.

\begin{figure*}[ht]
    \begin{minipage}{0.59\textwidth}
        \centering
        \hspace{-20pt}
        \begin{subfigure}[b]{0.46\textwidth}
            \includegraphics[width=\linewidth]{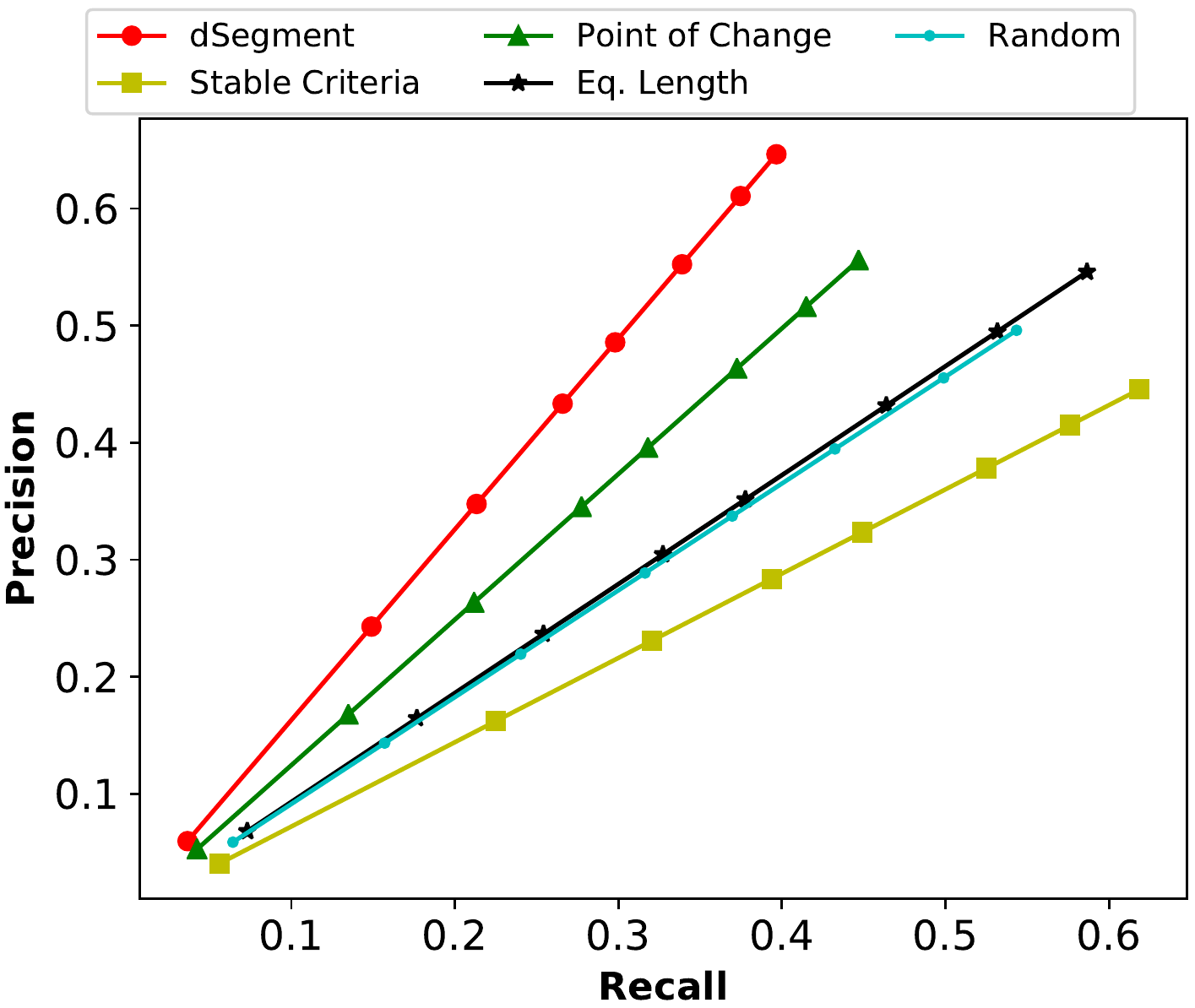}
            \caption{\scriptsize Comparison based on Easy-Aggregation set}
            \label{fig:pr_easy}
        \end{subfigure}\hspace{7pt}
        \begin{subfigure}[b]{0.46\textwidth}
            \includegraphics[width=\linewidth]{Precision-Recall.pdf}
            \caption{\scriptsize Comparison based on Strict-Aggregation set}
            \label{fig:pr_strict}
        \end{subfigure}
        \vspace{-5pt}
       \caption{\small Comparing different segmentation approaches based on DACT.}
    \end{minipage}\hfill
    \begin {minipage}{0.38\textwidth}
        \hspace{-20pt}
        \centering
        \includegraphics[width=\linewidth]{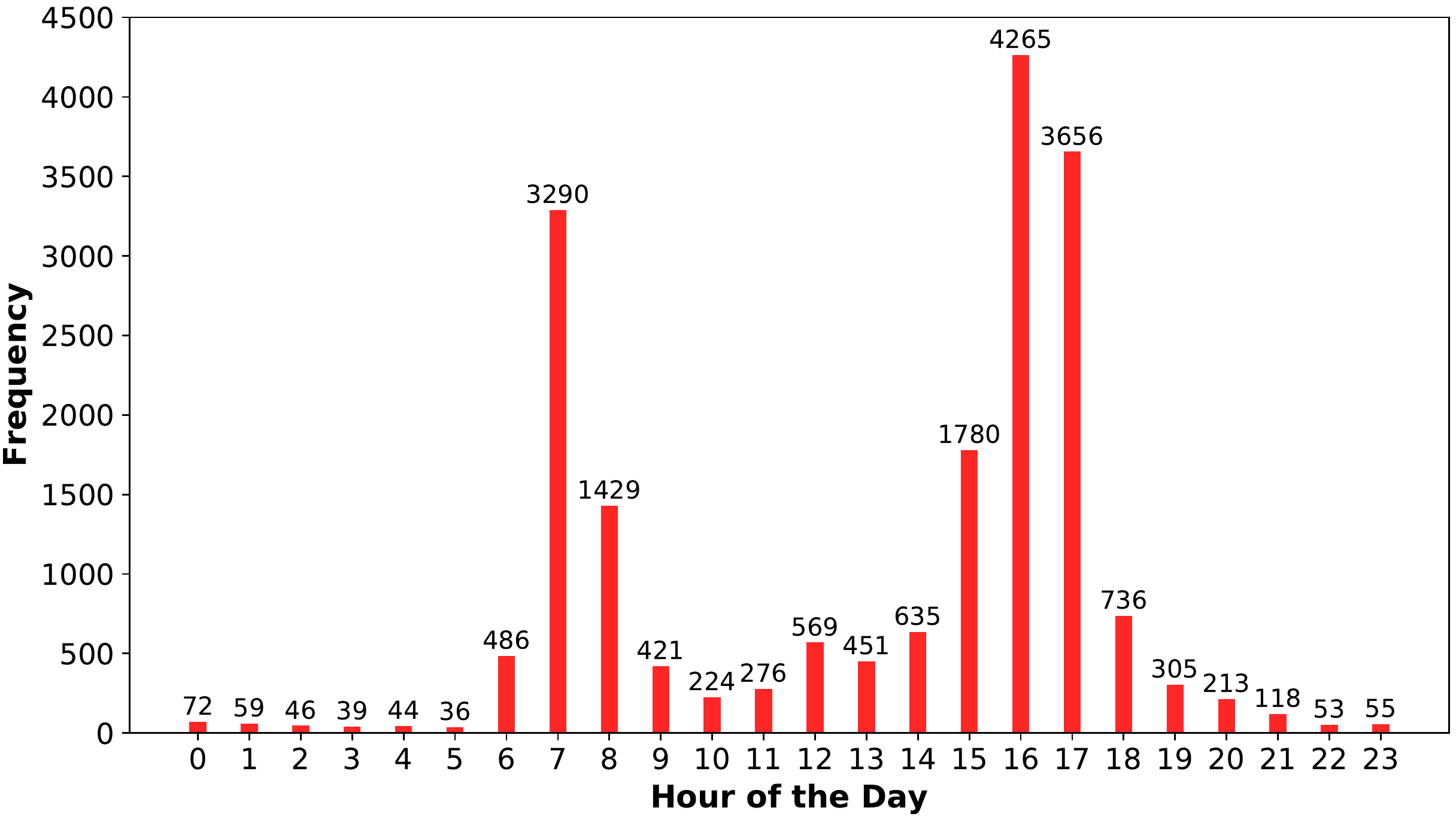}
        \vspace{-8pt}
        \caption{\small One year (from Feb 2016 to Feb 2017) congestion frequency distribution for Columbus Ohio based on Map Quest traffic reports.}
        \label{fig:OneYearColumbusCongestions_MQ}
    \end{minipage}
\end{figure*}

In order to find the upper bound on the number of existing segments, i.e., $K$ (see \cite{moosavi2016discovery}), we set $K=\frac{N}{5}$, where $N$ is the length of the trajectory. The minimum length of a segment is assumed to be $5$ (see \cite{moosavi2017dact}). Since we have two sets of annotations for trajectories in DACT, we evaluate and compare our approach based on both sets. 
Also, we use {\em Precision} and {\em Recall} as evaluation metrics. Given a trajectory $t$ with $n$ annotations $Ant_t = \langle a_1, a_2, \dots, a_n \rangle$, if algorithm $Alg$ finds $m$ cutting points $CP_{Alg} = \langle p_1, p_2, \dots, p_m \rangle$ for $t$, then we define precision and recall as follows:

\vspace{0.06in}
\hspace{0.2in} $Precision = \frac{Ant_t \hspace{0.03in} \cap \hspace{0.03in} CP_{Alg}}{m}$ \hspace{0.18in}  $Recall = \frac{Ant_t \hspace{0.03in} \cap \hspace{0.03in} CP_{Alg}}{n}$
\vspace{0.06in}

The intersection between $Ant_t$ and $CP_{Alg}$ is calculated as follows: to find a match for $p_i \in CP_{Alg}$, we calculate its Haversine distance to all {\em available} annotations in $Ant_t$. If we find a pair $(p_i, a_j)$, $a_j \in Ant_t$ for $1 \leq j \leq n$, such that their distance is lower than a pre-specified threshold, then we say there is a match for $p_i$. Once we find such an $a_j$, we no longer use it to match other cutting points in $CP_{Alg}$. 
We use values in set $\{0, 25, 50, 75, 100, 150, 200, 250\}$ as distance thresholds (in meters).
Figures \ref{fig:pr_easy} and \ref{fig:pr_strict} show the comparison between different segmentation approaches using gold sets in DACT.
For Equal Length and Random baselines, we use values $30$ and $50$ for $\eta$ based on Easy and Strict aggregation annotation sets. These numbers are set based on average number of segments in a trajectory, as reported in \cite{moosavi2017dact}.

The results show that {\sc dSegment} outperforms the other baselines by reasonable margins, based on both ground truth datasets. 
The maximum distance threshold (250 m) is obtained by dividing the average length of routes in the evaluation set (i.e., 10 km, see Table \ref{tab:segmentation_result}) by the average number of segments for a trajectory (i.e., 40\footnote{\scriptsize We have the average number of trajectories for {\em easy} and {\em strict} sets as 30 and 50, respectively \cite{moosavi2017dact}.}). 
Note that a solution which maximizes the precision is preferred, because we need {\em valid} segments to conduct a precise causality analysis to confidently derive the characteristics for a context.

%% file: CausalityResults.tex
\label{causality_analysis}
We define ``context'' as the combination of {\em location} (listed in Table~\ref{tab:segmentation_result}) and {\em time} (e.g., weekdays between 3pm to 7pm). 
We use two granularity levels for time: {\em Type of Day} (Weekday (WD) versus Weekend (WE)), and {\em Time of the Day} with five time intervals, i.e., {\em P1}: from 6am to 9:59am, {\em P2}: from 10am to 2:59pm, {\em P3}: from 3pm to 6:59pm, {\em P4}: from 7pm to 9:59pm, and {\em P5}: from 10pm to 5:59am. We used the one-year traffic congestion reports by Map Quest for the city of Columbus Ohio (see Figure \ref{fig:OneYearColumbusCongestions_MQ}) to derive aforementioned intervals. 

Table \ref{tab:segmentation_result} provides statistics on applying {\sc dSegment} on the test set, where the total number of extracted segments is 6,674.
{\sc dDescribe} can then be applied on this set to discover properties for each context. For that, we build an event database ${\mathcal E}$ from physical facts and temporal-physical events. We conduct the causality analysis by introducing a new correlation measure: suppose that for a set of trajectories $\mathcal{T} = \{T_1, T_2, \dots, T_M\}$ in context $C$, a sequence of cutting points $CP_i = \langle p_{i_1}, p_{i_2}, \dots \rangle$ for each $T_i \in \mathcal{T}$ is reported. Given an event database ${\mathcal E}$, we use Equation \ref{eqn:cvrg} to obtain the correlation for context $C$. 

\vspace{-0.2in}
\begin{equation}\label{eqn:cvrg}
    \small \mathit{Correlation}(C, {\mathcal E}) = \frac{\sum\limits_{i=1}^{i=M}\sum\limits_{p \in CP_i} \mathit{CheckRelevancy}(p,{\mathcal E})}{\sum\limits_{i=1}^{i=M} |CP_i|}
\end{equation}

In Equation \ref{eqn:cvrg}, $\mathit{CheckRelevancy}$ is a Boolean function depending on the type of event, as discussed below. 

\begin{figure*}[ht]
    \centering
    \hspace{-0.07\textwidth}
    \begin{subfigure}[b]{0.31\textwidth}
            \includegraphics[width=\linewidth]{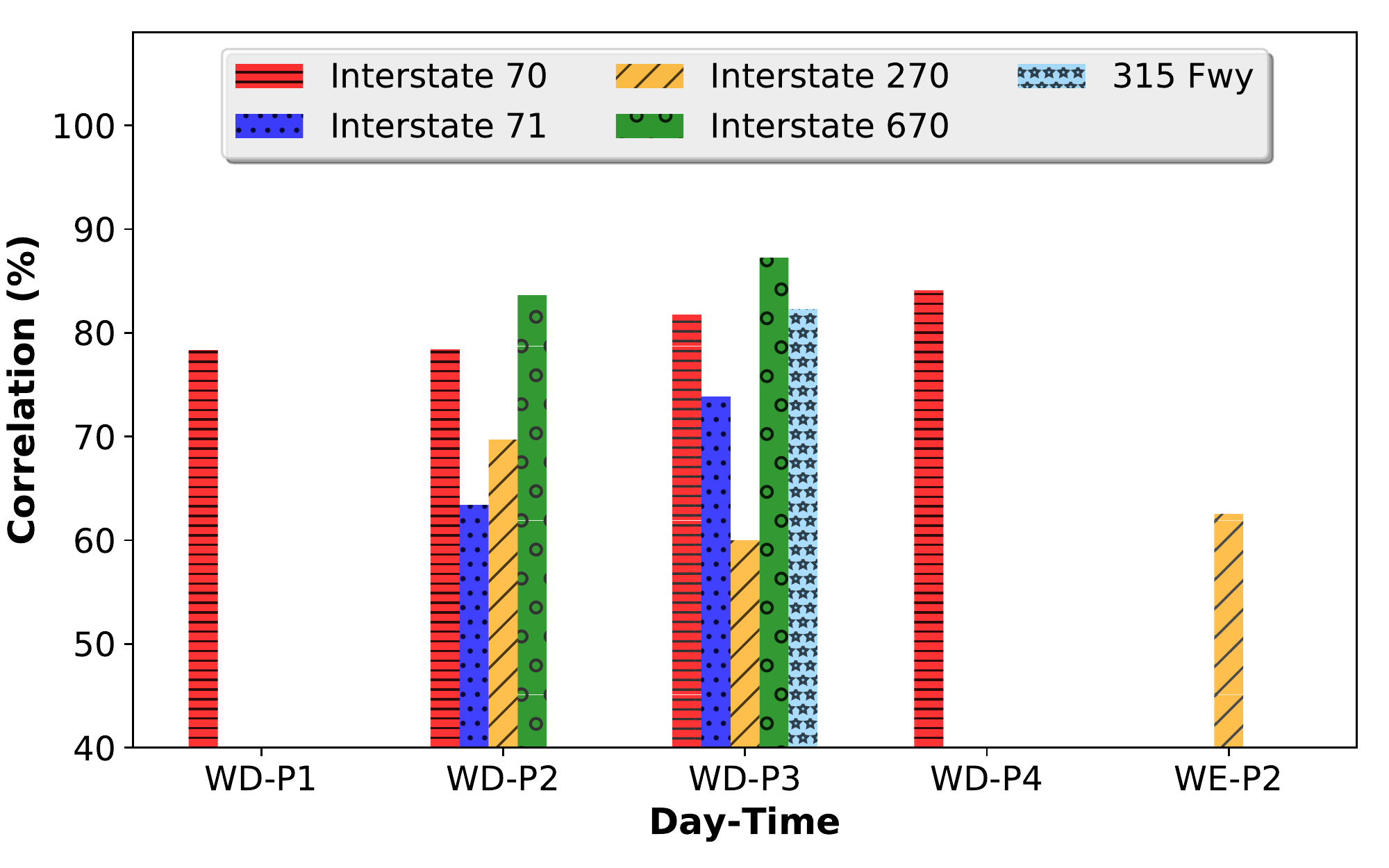}
            \caption{\scriptsize Physical}
            \label{fig:physical-facts}
    \end{subfigure}\hspace{0.0025\textwidth}
    \begin{subfigure}[b]{0.31\textwidth}
            \includegraphics[width=\linewidth]{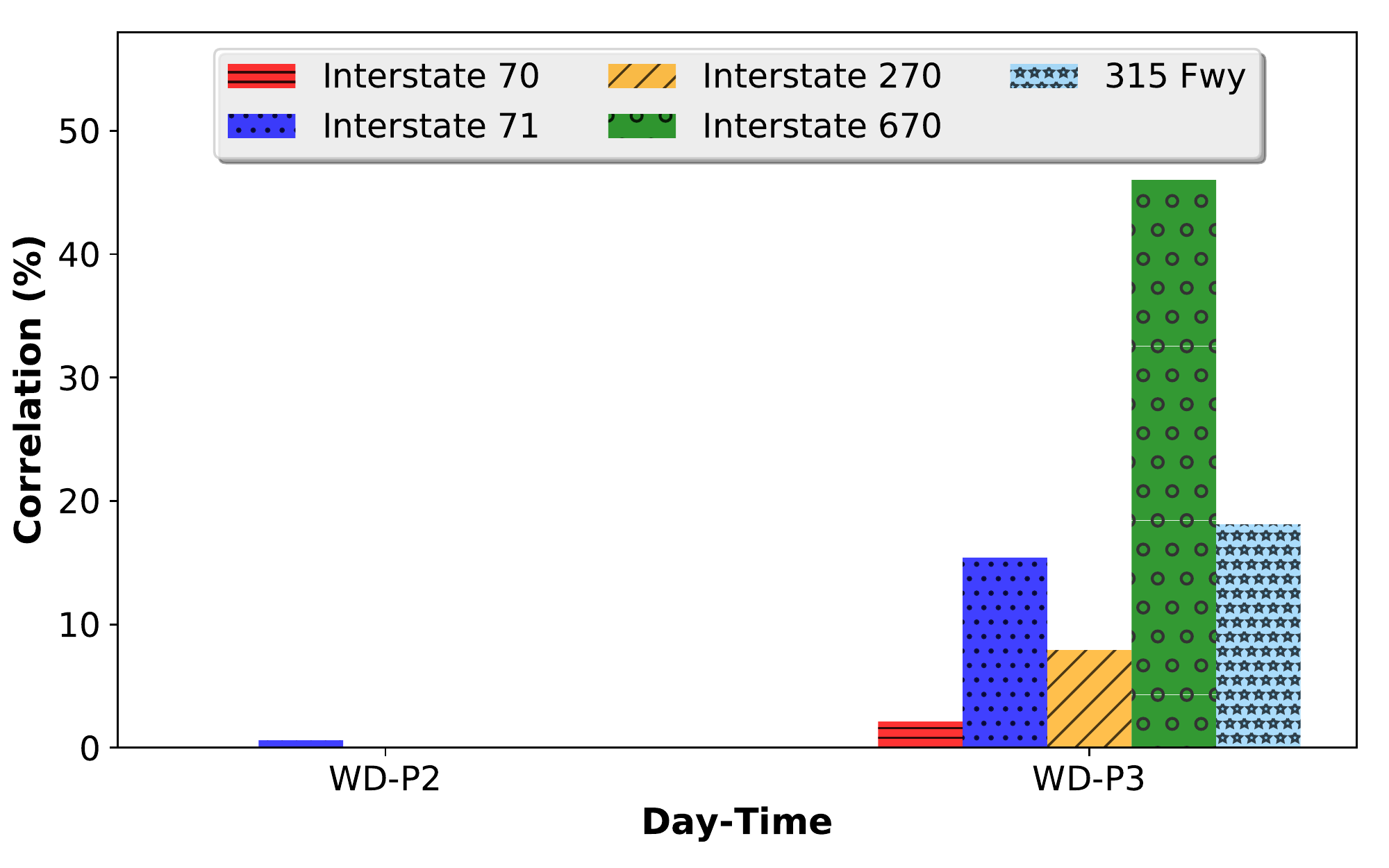}
            \caption{\scriptsize Temporal-Physical}
            \label{fig:temporal-physical}
    \end{subfigure}\hspace{0.0025\textwidth}
    \begin{subfigure}[b]{0.31\textwidth}
            \includegraphics[width=\linewidth]{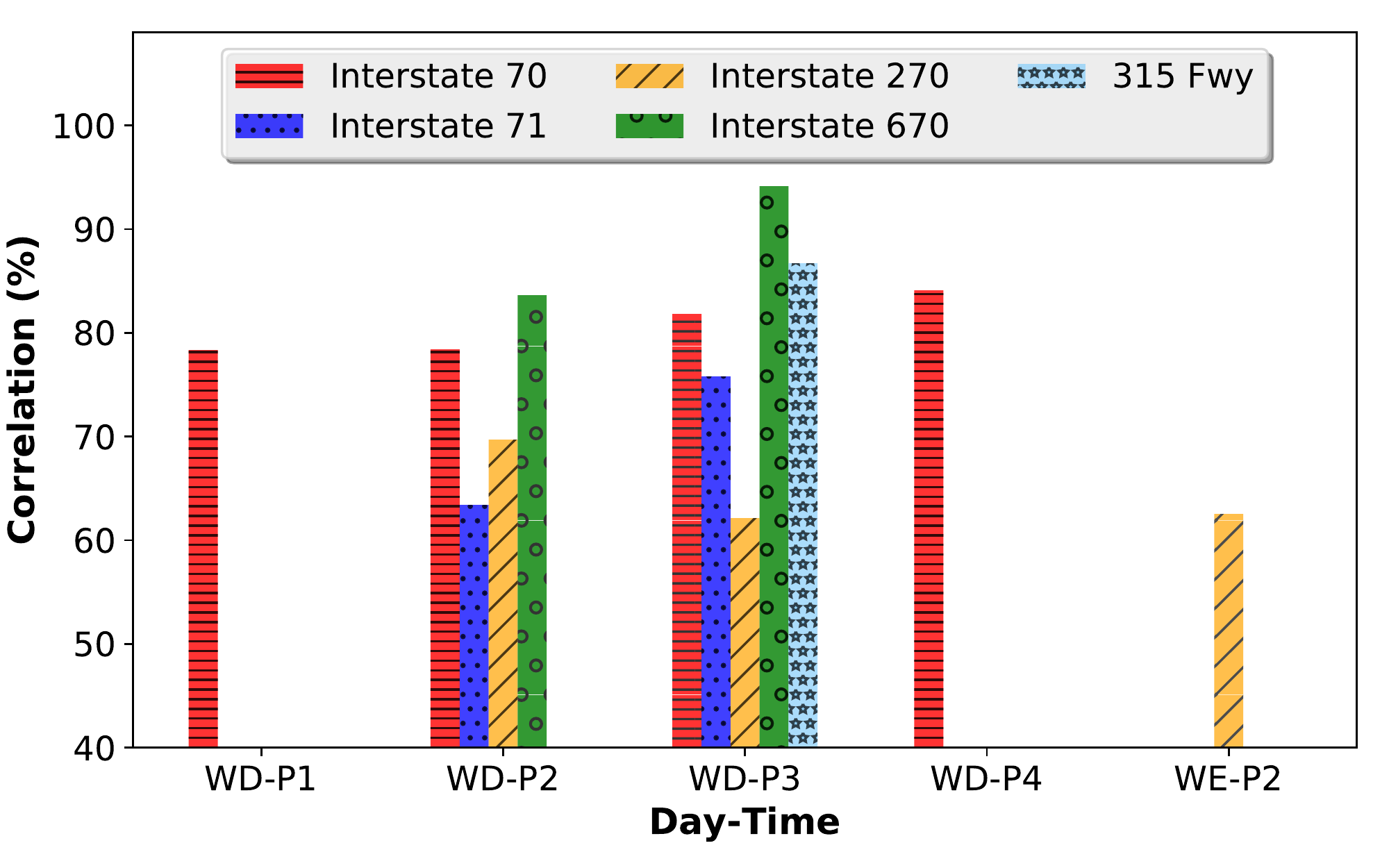}
            \caption{\scriptsize All}
            \label{fig:all}
    \end{subfigure}\hspace{-0.05\textwidth}
    \vspace{-0.14in}
    \caption{\small Correlation of extracted driving patterns (segments) with (a) Physical Facts, (b) Temporal-Physical Events, and (c) All Facts and Events.}
    \label{fig:correlation-study}
    \vspace{-0.08in}
\end{figure*}

\vspace{2pt}
\noindent\textbf{Event Data as Physical Facts}: First, we use physical facts to build the event database ${\mathcal E}$.
For this case, we define the $\mathit{CheckRelevancy}$ function by calculating the Haversine distance between a cutting point $p$ and a physical event $e \in \mathcal E$, and then checking if their distance is lower than a pre-specified threshold $th$ (empirically set to $200m$).
Figure \ref{fig:physical-facts} shows the correlation between the extracted segments of different contexts and physical facts. 
Note that the correlation analysis is only done for those contexts for which enough data exists.
We observe that on average, about 76.5\% of the driving patterns (segments) are correlated with the physical properties of routes.

\vspace{2pt}
\noindent\textbf{Event Data as Temporal-Physical Events}: We also employ temporal-physical events to build ${\mathcal E}$.
To define $\mathit{CheckRelevancy}$ in this case, we use the following two-step method:

\vspace{2pt}
\noindent{\em Step 1: Find potential congestion evidences.} Given a trajectory $T$, we find sub-trajectories of minimum length $5$ (based on \cite{moosavi2017dact}), where the speed of all points in such sub-trajectories is less than $55km/h$ (i.e., the average congestion speed in the traffic congestion dataset). We consider such sub-trajectories as showing potential evidence of congestion.

\vspace{2pt}
\noindent{\em Step 2: Finalization.} After finding potential evidence $\mathcal C$ of congestion, we scan through our traffic congestion dataset to see if there are at least 12 instances (i.e., one report per month) in the neighborhood (i.e., $200m$) of $\mathcal C$'s location, within the same day of the week and hour of the day (e.g., Tuesday 4pm).

\vspace{2pt}
We identify 465 traffic congestion sub-trajectories within 1,421 trajectories of the test set. The function $\mathit{CheckRelevancy}$ returns $1$ if a cutting point $p$ is in the neighborhood (i.e., $th=200m$) of at least one of traffic congestion evidences of a the trajectory $T$. Otherwise, it returns $0$. Figure \ref{fig:temporal-physical} illustrates the correlation analysis results between driving patterns and temporal-physical events. On average, \textasciitilde10.5\% of driving patterns were correlated with traffic congestion. 

\vspace{2pt}
\noindent\textbf{Event Data as Union of Fact and Events:} Finally, we consider both physical facts and temporal-physical events to build $\mathcal E$. In this case, $\mathit{CheckRelevancy}$ will function with respect to the type of the event. Figure \ref{fig:all} demonstrates the correlation of driving patterns with the set of all existing events. We observe that \textasciitilde78.1\% of segments are correlated with at least one of the event types. 
Moreover, side-by-side comparison of the results in Figures \ref{fig:temporal-physical} and \ref{fig:all}, reveals that both analyses lead to almost the same correlation patterns. This confirms that a significant number of segments are correlated with both sources of events.

\vspace{2pt}

Results from causality analysis are strong signals that capture the characteristics of a driving context. These insights can be employed in various applications. One example is usage-based insurance to study the behavior of an individual driver in order to evaluate how risky or safe he/she is, regarding the characteristics of context. Insights from our framework may also be used for driver coaching, by recommending further training to those drivers whose driving behavior in a context is not compatible with the properties of that context.


%% file: literature.tex
\vspace{-0.05in}
\section{Related Work}
\label{sec:rel}
Our study relates to research in {\em trajectory segmentation} (as in {\sc dSegment}) and {\em making sense of trajectories} (as in {\sc dDescribe}).

\vspace{2pt}
\noindent\textbf{Trajectory Segmentation:} The task of segmentation has been addressed in the literature in several studies. In \cite{buchin2010algorithmic}, a greedy segmentation algorithm exploits a set of monotonic spatiotemporal criteria (e.g., defining relative thresholds for some feature values) on features like speed, heading, etc. Alewijnse et al. extended this work to both monotonic and non-monotonic criteria \cite{alewijnse2014framework}. However, criteria-based methods need human input for tuning parameters.
Transforming trajectory prior to segmentation has also been previously discussed in \cite{panagiotakis2012segmentation}; however, their transformation is a local approach, based on comparing line segments of an input trajectory. On the other hand, we perform a global, likelihood-based transformation to provide a segmentation where the extracted segments represent meaningful driving patterns. 

\vspace{2pt}
\noindent\textbf{Making Sense of Trajectories.} Akin to {\sc dDescribe}, there are some other approaches which make sense of driving data and explore insights encapsulated in trajectories \cite{stenneth2011transportation,lou2009map,su2015making}. 
Wu et al. \cite{wu2016interpreting} predict traffic based on analysis of external data sources including Point Of Interest (POI) data, collision data, weather data, and geo-tagged tweet data. This work is similar to {\sc dDescribe}, where we discover correlations between driving patterns and traffic congestion and physical properties of routes.
However, we pursue a different goal which is the identification of characteristics of a context.

%% file: conc.tex
\vspace{-5pt}
\section{Conclusion}
\label{sec:conc}
We present \sys, a framework to derive characteristics of a context by extracting meaningful driving patterns ({\sc dSegment}), and then analyzing the extracted patterns ({\sc dDescribe}) to derive characteristics.
Our analysis shows how the {\sc dSegment} compares with the state-of-the-art in finding meaningful driving patterns. In addition, the results of {\sc dDescribe} show the ability of framework to interpret driving patterns which lead to new insights. Our future course of action is to incorporate more sources of event data,  such as Twitter feeds, in to {\sc dDescribe}.